\pdfoutput=1  %
\documentclass[letterpaper,journal]{IEEEtran}
\usepackage{amsmath,amsfonts,amssymb}
\usepackage[linesnumbered,ruled,vlined,algo2e]{algorithm2e}
\usepackage{array}
\usepackage[caption=false,font=footnotesize]{subfig}
\usepackage{textcomp}
\usepackage{stfloats}
\usepackage{placeins}
\usepackage{url}
\usepackage{verbatim}
\usepackage{graphicx}
\usepackage[numbers,sort&compress]{natbib}
\usepackage{booktabs}
\usepackage{multirow}
\usepackage{tabularx}
\graphicspath{{figures/}}
\usepackage[colorlinks=true,linkcolor=blue,citecolor=blue,urlcolor=blue,linktocpage=true]{hyperref}

\makeatletter
\let\ps@IEEEtitlepagestyle\ps@plain
\makeatother
\begin{document}

\title{Beyond Planning: Differentiable Control-in-the-loop End-to-End Autonomous Driving}

\author{Zhidong Wang, Zirui Li, Zitong Shan, Zhiqi Mao, Qihang Huang,\\ Zhan Chen, Jingsong Liang, Han Yu, and Chen~Lv
\thanks{Zhidong Wang, Zirui Li, Zitong Shan, Zhiqi Mao, Qihang Huang, Zhan Chen, Jingsong Liang, and Chen Lv are with the School of Mechanical and Aerospace Engineering, Nanyang Technological University, Singapore.
Zhidong Wang is also with the Collaborative Initiative, Interdisciplinary Graduate Programme, Nanyang Technological 
University, Singapore.
Han Yu is with the College of Computing and Data Science, Nanyang Technological University, Singapore.}
\thanks{Zhidong Wang and Zirui Li contributed equally to this work. Corresponding author: Chen Lv.}}

\maketitle

\begin{abstract}
End-to-end autonomous driving has achieved remarkable progress by planning trajectories directly from raw sensor data. Yet the step from planning to control still relies on hand-tuned tracking controllers that block the gradient path from vehicle motion back to the network. This gap stems from non-differentiable vehicle dynamics, aliased supervision from coarsely sampled expert labels, rate mismatch between high-rate control and low-rate re-planning, and vehicle model discrepancies between training and deployment. This paper proposes DIVA (\underline{D}ifferentiable \underline{I}dentified \underline{V}ehicle dynamics for end-to-end \underline{A}utonomous driving) to close this gap. An identified actuator-level vehicle model is embedded in the training loop of the end-to-end driving network, forming a differentiable chain from perception to vehicle states. A learning-based control decoder replaces the external controller and is supervised on the rolled-out vehicle states, bypassing the aliased control-input labels. A rate-consistent control-input parameterization reconciles the heterogeneous sampling rates and enables asynchronous execution with a 3.8\,ms control-only cycle. On-policy re-identification and an online disturbance observer address training-time distribution shift and deployment-time model mismatch respectively. On the closed-loop Bench2Drive benchmark, DIVA achieves a driving score of 86.7, surpassing the widely adopted state-of-the-art approach by 3.3 points while nearly halving mean jerk and reducing brake switching by 64\%, and maintains superior path tracking accuracy under significant vehicle model mismatch at deployment.
\end{abstract}


\section{Introduction}
\label{sec:intro}
\IEEEPARstart{E}{nd}-to-end autonomous driving unifies perception and planning within a single differentiable model that maps raw sensor data to a reference trajectory~\cite{transfuser,uniad}. Propagating the planning losses back through the perception backbone reduces the compound error that accumulates across handcrafted module boundaries. The planned trajectory serves as the imitation target and is tracked by lateral and longitudinal controllers at deployment. Training thus stops at the planning output, while closed-loop performance depends on the actual control response of the vehicle: the tracking controller, the actuator subsystems and the chassis dynamics remain outside the differentiable pipeline and block the gradient path from the vehicle response back to the network.

The execution of the end-to-end planning results relies on classical controllers designed for modular pipelines, which assume a smooth, feasible trajectory and a known vehicle model~\cite{thrun2006stanley,paden2016survey}. An end-to-end planner satisfies neither assumption: its output is temporally inconsistent, exhibits discrete speed jumps and loses reliability over long horizons. The resulting planning-to-control discontinuity comprises three gaps. The \emph{supervision gap}: expert commands are recorded at a label rate far below the control rate, yielding aliased control-input labels. The \emph{gradient gap}: the non-differentiable controller and vehicle dynamics keep the training loss from reflecting the actual control response. The \emph{deployment gap}: the vehicle model presumed in training differs from the deployed vehicle, and the label, perception and control rates also differ.

\begin{figure*}[!t]
\centering
\includegraphics[width=\textwidth]{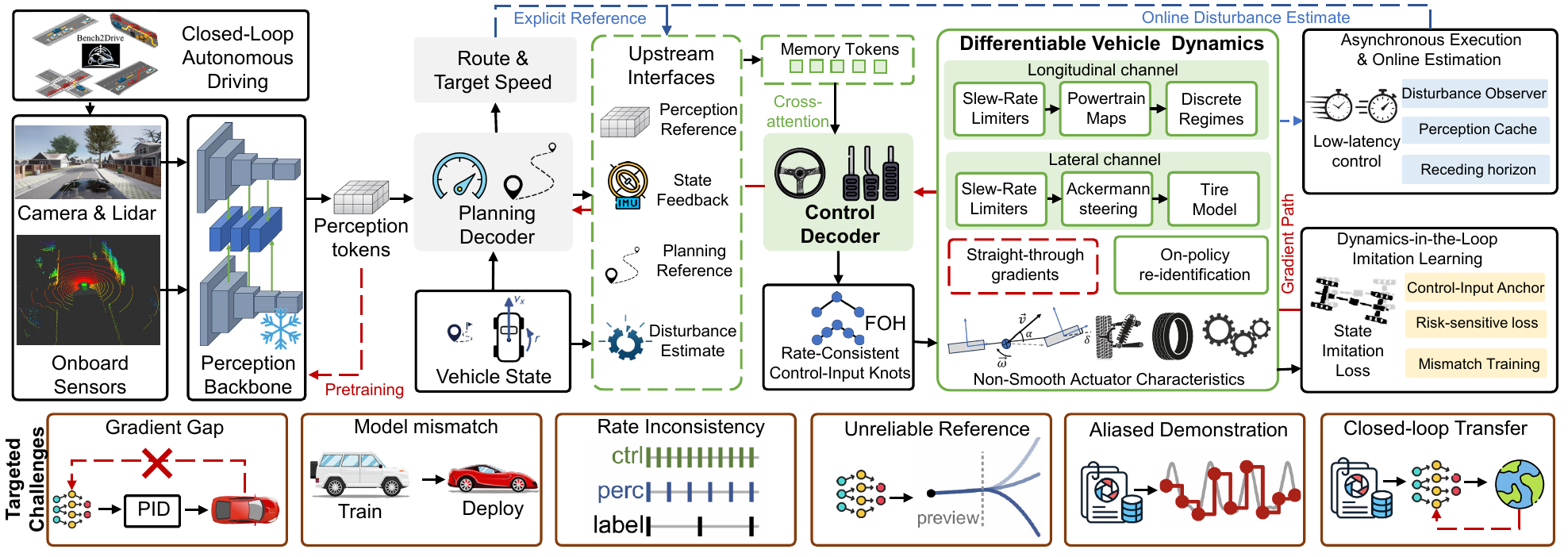}
\caption{Overview of DIVA: the control decoder cross-attends to memory tokens that combine the scene and planning features, the explicit route and target-speed reference, the state feedback and the disturbance estimate, and its control-input knots pass through a first-order hold into the identified differentiable vehicle dynamics, whose longitudinal and lateral actuator channels drive a single-track model integrated at 20\,Hz. Under imitation learning in state space, the state loss propagates along the gradient path back through the vehicle dynamics into the control and planning decoders, and at deployment the control decoder runs on a fast path over cached perception, executing receding-horizon control inputs conditioned on the disturbance estimate of the online observer.}
\label{fig:overview}
\end{figure*}

This paper proposes \textbf{DIVA} (\textbf{D}ifferentiable \textbf{I}dentified \textbf{V}ehicle dynamics for end-to-end \textbf{A}utonomous driving) to bridge these gaps. DIVA embeds a differentiable actuator-level vehicle dynamics model in the training loop, forming a single differentiable pathway from perception to vehicle states (Fig.~\ref{fig:overview}), and replaces the external controller with a learned control decoder whose supervision, gradient flow and deployment all pass through the identified vehicle dynamics model. The main contributions are:
\begin{itemize}
    \item \textbf{End-to-end autonomous driving model differentiable through to vehicle states.} An  actuator-level vehicle dynamics model links the planning output to the vehicle response, making the architecture differentiable from perception to vehicle states; on-policy re-identification keeps the model consistent with the learned controller.
    \item \textbf{Dynamics-in-the-loop imitation learning in state space.} Supervision on vehicle states through the dynamics model bypasses the aliased control-input labels that make direct control-input cloning ill-posed; an identifiability-aware loss and a gradient gate enable joint training of the controller and planner.
    \item \textbf{Rate-consistent asynchronous execution at deployment.} A rate-consistent control-input parameterization reconciles the label, perception and control rates; asynchronous execution decouples control from perception latency, enabling full-rate feedback and closed-loop robustness under vehicle model mismatch.
\end{itemize}

\section{Related Work}
\label{sec:related}
\subsection{The Planning-to-Control Interface in Autonomous Driving }
\label{subsec:rw-interface}
In modular stacks, planning interfaces with control through a reference trajectory tracked by geometric, PID, linear-quadratic or model predictive controllers~\cite{stano2023model,artunedo2024lateral}, with modern MPC designs incorporating actuator delays and tire limits~\cite{lee2023robust}. In end-to-end stacks, the output representation determines where this interface resides~\cite{chen2024end}: end-to-end networks output control commands~\cite{yang2026raw2drive}, trajectory--control fusions~\cite{wu2022trajectory}, temporal trajectories~\cite{jiang2023vad,jia2025drivetransformer}, or geometric paths to be tracked~\cite{renz2025simlingo}, where retuning controller gains alone can improve driving performance significantly~\cite{liang2024tuning}. Yet the training always stops at the planning output, and the learned reference trajectory is state-conditioned, temporally inconsistent and reliable only within the preview distance.

Several works extend the differentiable chain beyond the planner through kinematic or nominal vehicle models: DiffStack~\cite{karkus2023diffstack} employs a unicycle-based LQR, analytic policy gradients~\cite{nachkov2025autonomous,nachkov2026unlocking} supervise through bicycle dynamics, Rodr\'iguez-Vidal \emph{et al.}~\cite{rodriguez2026addressing} enable waypoint-space supervision via a kinematic model, and DIPC~\cite{zhang2026dynamics} couples a differentiable planner with a neural MPC through a six-state vehicle model with Magic Formula tires. In all cases, the vehicle model is kinematic or nominal with assumed parameters, without actuator dynamics such as slew-rate limits or lags, and the label, perception and control rates are not decoupled.

\subsection{Learning-based Controllers Considering Model Mismatch}
\label{subsec:rw-mismatch}
Reinforcement learning~\cite{liu2024reinforcement} and imitation learning~\cite{gong2024beyond} can train end-to-end vehicle controllers, but the learned dynamics are implicitly encoded in the network weights and cannot be re-calibrated when the vehicle dynamics changes. The classical remedy pairs a nominal model with an observer: unknown-input observers generate steering commands~\cite{tan2024twin}, and adaptive Kalman filters supply tire and friction estimates to hand-designed controllers~\cite{qin2022lateral}. Hybrid neural--physical structures increasingly identify the nominal model itself~\cite{zhou2024vehicle,wu2024path}. On the learning side, domain randomization tolerates a range of vehicle dynamics characters~\cite{ghignone2024tc}, and continual meta-learning adapts dynamics models online~\cite{tsuchiya2024online}. Explicit disturbance estimates have been supplied to learned quadrotor controllers~\cite{huang2023datt}, while for ground vehicles online adaptation refines the dynamics model within an optimizer~\cite{djeumou2024one}; DIPC examined certain vehicle parameter variation without retraining or adaptation~\cite{zhang2026dynamics}.

These approaches remain disconnected: observer-based methods serve rule-based controllers, learned controllers absorb mismatch through domain randomization, and online adaptation refines the dynamics model rather than the controller. No existing work feeds explicit mismatch estimates to a learned controller within an end-to-end driving stack; DIVA does so by unifying model identification, differentiable planning and mismatch-aware control in a single trainable framework.

\section{Methodology}
\label{sec:method}

\subsection{Overall Framework}
\label{subsec:framework}

The end-to-end autonomous driving networks are supervised at the planning level; DIVA extends the differentiable computation graph to the vehicle-state level so that the training loss directly reflects closed-loop execution performance.

\subsubsection{Architecture Overview}
\label{subsubsec:architecture}
Fig.~\ref{fig:overview} illustrates the overall architecture. A perception backbone fuses camera and LiDAR observations into perception tokens, from which a planning decoder produces a reference route and a target speed distribution. A subsequent control decoder maps these planning outputs together with state feedback and disturbance estimate into control-input knots that are upsampled to the control rate via first-order hold (FOH) interpolation. The identified vehicle dynamics model then propagates these control inputs forward over the prediction horizon, yielding predicted vehicle states that are supervised against the expert demonstrations. The control decoder's outputs influence the training loss exclusively through the differentiable vehicle model: the state-loss gradient back-propagates through the vehicle dynamics into the control decoder and, modulated by a learnable gradient gate, further into the planning decoder. At deployment, the control decoder executes at every control step with cached perception features and receives an online disturbance estimate, enabling real-time mismatch-aware control.

\subsubsection{Problem Formulation}
\label{subsubsec:problem}
The planner output is a geometric path decoupled from a target speed~\cite{transfuser++}, which performs markedly better in closed-loop driving than a temporal trajectory~\cite{uniad}. It consists of a route $\mathcal{R}=(\boldsymbol{\rho}_1,\ldots,\boldsymbol{\rho}_N)$ of $N$ points $\boldsymbol{\rho}_i\in\mathbb{R}^2$ in the vehicle frame, uniformly spaced along the arc length, and a probability vector $\mathbf{p}^v$ over discrete speed classes. The vehicle accepts a actuator command vector $\mathbf{c}=[c_t,c_b,c_s]^\top$ of throttle, brake and steering at the control rate $f_{\mathrm{ctrl}}=20$\,Hz, with control period $T_s=1/f_{\mathrm{ctrl}}$. The expert demonstrations are recorded at $f_{\mathrm{lab}}=4$\,Hz with label period $T_{\mathrm{lab}}=1/f_{\mathrm{lab}}$: for a frame at time $t$, the labels span a 2\,s horizon and comprise $K=8$ future expert states $\mathbf{z}^\star_j=[X^\star_j,Y^\star_j,\psi^\star_j,v^\star_j]^\top$ of position, heading and speed at $t+jT_{\mathrm{lab}}$, $j=1,\ldots,K$, and the recorded expert commands, expressed as control inputs $\mathbf{u}^\star_j$, $j=0,\ldots,K$. Perception runs at a rate $f_{\mathrm{perc}}\le f_{\mathrm{ctrl}}$, bounded by the latency of the backbone and the planner.

The control decoder $\mathcal{C}$ maps the memory tokens to $K+1$ control-input knots $\mathbf{U}$ at the label rate. The knots are interpolated to a control-rate sequence $\bar{\mathbf{u}}_{0:H-1}$ over $H=5K$ control steps,  and rolled out through the identified vehicle dynamics model $\mathbf{z}_{k+1}=\mathcal{F}\left(\mathbf{z}_k,\mathbf{c}(\bar{\mathbf{u}}_k);\boldsymbol{\theta}\right)$, written as $\mathbf{z}_{1:H}=\Phi\left(\mathbf{z}_0,\bar{\mathbf{u}}_{0:H-1};\boldsymbol{\theta}\right)$, where $\mathbf{z}$ is the model state and $\boldsymbol{\theta}$ collects vehicle dynamics mismatch parameters, drawn from a distribution $p(\boldsymbol{\theta})$ in training. The map $\mathbf{c}(\cdot)$ from control inputs to commands is given in Section~\ref{subsubsec:three-clocks}. The control decoder is trained to solve
\begin{equation}
\min_{\mathcal{C}}\;\mathbb{E}_{\mathcal{D},\,\boldsymbol{\theta}\sim p(\boldsymbol{\theta})}\Big[\sum_{j=1}^{K}\ell\left(\mathbf{z}_{5j},\mathbf{z}^\star_j\right)\Big],
\label{eq:problem}
\end{equation}
where $\mathcal{D}$ denotes the demonstrations and $\ell$ is the discrepancy between $\mathbf{z}_{5j}$ and $\mathbf{z}^\star_j$ in position, heading and speed. Eq.~\eqref{eq:problem} addresses control rather than prediction: the control inputs are the unknowns, and the model maps them to the states that the demonstrations record. The planning decoder is trained jointly with its own loss.

\subsection{Control Decoder}
\label{subsec:controller}

The learned planner regenerates its reference at every perception update, conditioned on the current vehicle state and reliable only within the preview horizon. A controller that relies solely on this reference inherits its uncertainty. To mitigate this limitation, the control decoder receives not only the explicit reference but also the raw scene tokens and the planner's intermediate features, enabling compensation for planning inaccuracies through richer contextual information.

\subsubsection{Memory Tokens}
\label{subsubsec:decoder}
The perception backbone and the planning decoder follow TF++~\cite{transfuser++}. The fused bird's-eye-view feature is projected and positionally encoded into perception tokens $\mathbf{M}$, augmented by a speed and navigation token. A transformer with learned queries $\mathbf{Q}_p$ attends to $\mathbf{M}$ and produces planning tokens $\mathbf{P}$, from which a GRU decodes the route $\mathcal{R}$ and a softmax head outputs the speed distribution $\mathbf{p}^v$.

The control decoder is an independent three-layer transformer whose memory
\begin{equation}
\begin{aligned}
\mathbf{S}=\big[&\phi_M(\mathbf{M});\phi_P(\mathbf{P});E_R(\mathcal{R}_{\mathrm{ref}});\\
&E_v(\mathbf{p}^v_{\mathrm{ref}});E_o(\mathbf{o});E_d(\mathbf{d})\big]
\end{aligned}
\label{eq:memory}
\end{equation}
concatenates four groups of tokens, formed by learned adapters $\phi_M$, $\phi_P$ and learned embeddings $E_R$, $E_v$, $E_o$, $E_d$. The feature interface $\phi_M(\mathbf{M})$, $\phi_P(\mathbf{P})$ lets the controller attend to the scene and the planner's intermediate representation. The explicit reference, per-point route tokens $E_R(\mathcal{R}_{\mathrm{ref}})$ and a speed token $E_v(\mathbf{p}^v_{\mathrm{ref}})$, keeps the interpretable interface of the planner. The state-feedback token $E_o(\mathbf{o})$, with $\mathbf{o}=[v_x,r_{\mathrm{avg}}]^\top$, closes the loop on speed and yaw rate and deliberately excludes accelerations. The disturbance-estimate token $E_d(\mathbf{d})$ carries the observer estimates. Here $r_{\mathrm{avg}}$ is the yaw rate averaged over one label period $T_{\mathrm{lab}}$. At deployment, the reference $(\mathcal{R}_{\mathrm{ref}},\mathbf{p}^v_{\mathrm{ref}})$ is the planner output $(\mathcal{R},\mathbf{p}^v)$. In training, it is replaced by the ground truth with probability 0.5, and the feature tokens are masked with probability 0.1, so that the controller can also operate on the explicit reference alone.

\subsubsection{Rate-Consistent Control-Input Knots}
\label{subsubsec:three-clocks}
The decoder output yields the knots
\begin{equation}
\mathbf{U}=\tanh\left(\operatorname{MLP}\left(\operatorname{Dec}_c\left(\mathbf{Q}_c,\mathbf{S}\right)\right)\right)\in[-1,1]^{(K+1)\times2},
\label{eq:knots}
\end{equation}
where $\operatorname{Dec}_c$ denotes this transformer with $K+1$ learned queries $\mathbf{Q}_c$. Each knot $\mathbf{u}_j=[u_{x,j},u_{s,j}]^\top$ is converted to vehicle commands by $\mathbf{c}(\mathbf{u})=\left[[u_x]_+,[-u_x]_+,u_s\right]^\top$, where $[x]_+=\max\{x,0\}$. The single signed longitudinal input $u_x$ excludes simultaneous throttle and brake and keeps the longitudinal acceleration monotone in $u_x$. As absolute commands are bounded by $\tanh$, the knots require no previous-command input and cannot drift like an integrator.

Parameterizing the control inputs at the control rate would introduce far more free inputs per sample than the state labels can constrain, overfitting to the temporal aliasing. DIVA therefore places the knot values at the label rate, but integrates the model and executes the commands at the control rate. The knots are interpolated to the control period by FOH,
\begin{equation}
\bar{\mathbf{u}}_k=\left(1-\left\{\tfrac{k}{5}\right\}\right)\mathbf{u}_{\lfloor k/5\rfloor}+\left\{\tfrac{k}{5}\right\}\mathbf{u}_{\lfloor k/5\rfloor+1},
\label{eq:foh}
\end{equation}
for $k=0,\ldots,H-1$, where $\lfloor\cdot\rfloor$ and $\{\cdot\}$ denote the integer and fractional parts. The loss is evaluated only at the control steps that coincide with the label instants, with control step~0 applying knot~0 exactly. Retaining the full control-rate integration is essential: coarsening the model to one step per knot would introduce numerical distortion, and the controller would learn to compensate for this model bias rather than for the true vehicle dynamics. During deployment, the control decoder executes at every control step with the latest state feedback $\mathbf{o}$ and disturbance estimate $\mathbf{d}$, applying only knot~0 in a receding-horizon fashion. The knot parameterization therefore serves purely as a training-time mechanism for temporal credit assignment and does not reduce the closed-loop control rate.

\begin{figure}[!t]
\centering
\includegraphics[width=\columnwidth]{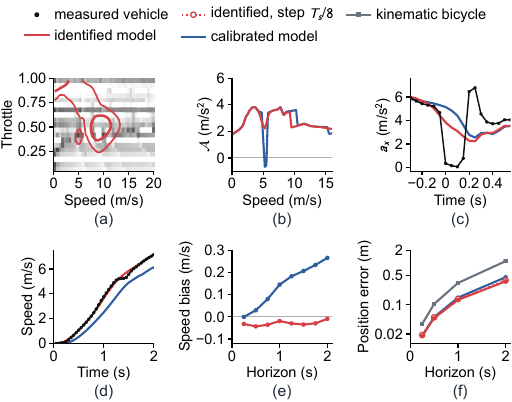}
\caption{On-policy re-identification: (a) coverage of the identification excitations (gray) and of the base controller (red, 50\% and 85\% contours); (b) map at throttle 0.7; (c) 141 on-policy upshifts; (d) 49 on-policy launches; (e) speed bias of both models under the controller's commands; (f) position error on held-out drives; kinematic bicycle: the constant-parameter model of~\cite{chen2021learning} shipped with TF++.}
\label{fig:onpolicy}
\end{figure}

\subsection{Identified Differentiable Vehicle Dynamics}
\label{subsec:dynamics}

The vehicle dynamics model $\mathcal{F}$ constitutes the sole differentiable path from the control decoder's output knots to the predicted vehicle states; consequently, the controller implicitly adapts its commands to the characteristics of this model. This coupling imposes three requirements on~$\mathcal{F}$. First, the model must faithfully capture the actuator dynamics that govern the vehicle's response to a given command. Second, it must provide well-defined gradients throughout the controller's operating envelope, including at standstill. Third, it must generalize beyond the identification excitations and remain accurate under the command distribution induced by the trained controller. In addition, $\mathcal{F}$ serves a fourth role by parameterizing the vehicle-side model mismatch for online estimation.

\subsubsection{Model Structure}
\label{subsubsec:model-structure}
The state of the model is
\begin{equation}
\mathbf{z}=\left[X,Y,\psi,v_x,v_y,r,\tilde c_s,\tilde c_t,\tilde c_b,a_d,a_b,\chi\right]^\top,
\label{eq:state}
\end{equation}
which comprises the vehicle pose $(X,Y,\psi)$, the longitudinal and lateral velocities $v_x$ and $v_y$, the yaw rate $r$, and the actuator states: the rate-limited steering, throttle and brake inputs $\tilde c_s$, $\tilde c_t$ and $\tilde c_b$, the lagged powertrain and brake accelerations $a_d$ and $a_b$, and a wheel-lock indicator $\chi\in\{0,1\}$.

The model explicitly represents the actuator chain from commanded input to vehicle response. Each command first passes through the vehicle's slew-rate limiter, which enforces the physical rate constraints on steering, throttle and brake. The powertrain acceleration $a_d$ is governed by a static map $\mathcal{A}(v_x,\tilde c_t)$, filtered through a first-order lag; this map encodes the steady-state acceleration on a flat road, accounting for road load, engine braking and the gear selected by the transmission shift schedule at the current speed. The brake acceleration $a_b$ is similarly modeled as a constant deceleration $a_{\mathrm{brk}}<0$:
\begin{equation}
a_{x,k}=a_{d,k+1}+a_{b,k+1}-g\sin\gamma_k,
\label{eq:longitudinal}
\end{equation}
where $g$ is the gravitational acceleration and $\gamma_k$ is the road grade angle, positive uphill. Two discrete regimes, a wheel lock in the first gear and a standstill hold under braking, complete the longitudinal channel. In the lateral channel, the steering input sets the road-wheel angle through a speed-dependent steering limit and the Ackermann geometry. A single-track model with saturating lateral tire forces yields $v_y$ and $r$, and blends into the kinematic single-track model at low speed.

The model advances by one control period $T_s$ per step and is formulated directly in discrete time rather than as a numerical discretization of continuous-time dynamics; empirically, reducing the integration step below $T_s$ yields no improvement in open-loop accuracy on held-out drives (Fig.~\ref{fig:onpolicy}(f)). This discrete-time formulation also motivates the design choice: the control-input knots are interpolated to the control rate and integrated at that rate, rather than integrated at the coarser label rate. The model parameters are identified through a three-tier procedure in which each tier is fitted with all preceding tiers held fixed: specification-level parameters are drawn from the vehicle datasheet, calibration-level parameters are obtained from constant-input, step and coast-down experiments, and the remaining parameters are identified from step-steer and swept-sine maneuvers. Appendix~\ref{app:parameters} provides the complete model equations, the identification protocol, the resulting parameter values, and a validation against measured vehicle data.

\subsubsection{Gradients Through Non-Smooth Actuator Characteristics}
\label{subsubsec:gradients}
Several elements of $\mathcal{F}$ are non-smooth: the split of $u_x$ into throttle and brake, the floor $[\cdot]_+$ on the speed, the dead band of $\mathcal{A}$, the discrete regimes, the slew-rate limiter and the clipping of the commands. A zero gradient there would stop learning exactly where the vehicle launches and stops. The split and the floor therefore use a leaky straight-through estimator~\cite{bengio2013estimating}: the forward pass computes $[x]_+$, and the backward pass uses the slope 1 for $x>0$ and $\epsilon_{\mathrm{st}}=0.1$ otherwise. Gradients thus still flow at standstill and in the dead band. The indicators of the discrete regimes are detached in the backward pass, and the limiter and the clipping keep the subgradient. The Ackermann relation is implemented in a form that is continuously differentiable at zero steering (Appendix~\ref{app:parameters}). A finite-difference check confirms the analytical gradients with respect to the knots.

\subsubsection{Policy-Induced Model Shift and On-Policy Re-Identification}
\label{subsubsec:onpolicy}
System identification relies on designed excitations, such as constant inputs, steps and bang-bang sequences, which hold a few discrete input levels (Fig.~\ref{fig:onpolicy}(a)). A controller trained through the model instead applies smooth, moderate throttle and modulated braking. It drives the model into regimes that the identification data cover only sparsely, and inherits the model error there. This paper terms the effect \emph{policy-induced model shift}, the model-side counterpart of the covariate shift of imitation learning; it instantiates identification for control, under which the identification experiment should reflect the operating conditions of the controller~\cite{gevers2005identification}.

Replaying open loop the closed-loop telemetry of a first controller trained through the calibrated model reveals a systematic speed over-prediction (Fig.~\ref{fig:onpolicy}(e)). The error stems from the upshift torque interruptions, the shift points and the brake-release lag (Fig.~\ref{fig:onpolicy}(b)--(d)). The correction has deliberately low capacity with 12 scalars in total: it edits only the tractive part of $\mathcal{A}$ and the two lag constants (Appendix~\ref{app:parameters}). It largely eliminates the speed bias under the first controller's commands and improves the accuracy on the held-out drives (Fig.~\ref{fig:onpolicy}(e) and (f)). DIVA is trained through the re-identified model.

\subsubsection{Mismatch Parameterization}
\label{subsubsec:observer}
The deployed vehicle never matches the training model exactly: the model structure is simplified, and the mass, actuator gains, biases and friction change in operation. These deviations are parameterized by
\begin{equation}
\boldsymbol{\theta}=\left[k_s,b_s,b_t,k_d,k_b,k_m,k_\mu,d_x\right]^\top.
\label{eq:theta}
\end{equation}
The steering gain $k_s$, the steering bias $b_s$ and the throttle bias $b_t$ act on the commands before the slew-rate limiter, $c_s\leftarrow k_sc_s+b_s$ and $c_t\leftarrow c_t+b_t$. The powertrain and brake gains $k_d$ and $k_b$ scale $a_d$ and $a_b$. The mass factor $k_m$ divides the longitudinal acceleration and scales the mass, the yaw moment of inertia and the cornering stiffnesses jointly, which leaves the lateral response unchanged. The tire--road friction factor $k_\mu$ scales the tire saturation and the locked-wheel deceleration, and $d_x$ adds a longitudinal acceleration disturbance. The acceleration in Eq.~\eqref{eq:longitudinal} thus becomes $a_{x,k}=(k_da_{d,k+1}+k_ba_{b,k+1})/k_m+d_x-g\sin\gamma_k$, and the nominal value $\boldsymbol{\theta}_0=[1,0,0,1,1,1,1,0]^\top$ recovers the identified model.

Given that only speed and yaw rate are available as measurements, the individual components of $\boldsymbol{\theta}$ acting on the same dynamic channel are not separately observable. The observable quantities reduce to two lumped deviations,
\begin{equation}
\begin{aligned}
\bar b_s&=b_s+\left(k_s-1\right)c_s,\\
\bar d_x&=d_x+\left(\frac{k_d}{k_m}-1\right)a_d+\left(\frac{k_b}{k_m}-1\right)a_b-g\sin\gamma,
\end{aligned}
\label{eq:lumped}
\end{equation}
representing the net steering bias and the total unexplained longitudinal acceleration. These two lumped deviations constitute the disturbance signals on which the control decoder is conditioned: during training they are computed analytically from $\boldsymbol{\theta}$, while at deployment they are estimated online. The identified model consequently fulfills three roles: it serves as the differentiable plant for controller training, as the process model of the online observer, and, through the observer's estimates, as the dynamics model underlying the classical baseline controllers.

\subsection{Dynamics-in-the-Loop Imitation Learning}
\label{subsec:learning}

With the differentiable chain from control decoder to vehicle states now defined, this section addresses the training procedure: the supervision signal and loss formulation, the strategies that ensure closed-loop validity and robustness to model mismatch, and the joint training of the controller with the planner. 

\subsubsection{State-Space Supervision With Identifiability-Aware Control-Input Anchor}
\label{subsubsec:supervision}
The design of the supervision signal is governed by two properties of the expert demonstrations. First, the expert commands are recorded at a rate lower than the control rate, so the raw control-input labels suffer from temporal aliasing and cannot faithfully reproduce the expert motion; in contrast, control-input knots recovered from the expert states via gradient-based inversion yield substantially more accurate trajectories. Second, the longitudinal command is not identifiable from the vehicle motion at standstill, since any command within the pedal dead band produces zero acceleration and leaves the vehicle state unchanged.

The supervision therefore combines a state-space loss with a control-input anchor, where the anchor is weighted more heavily in regimes where the state provides insufficient gradient signal (e.g., standstill) or where the control-input labels are reliable. Specifically, the state loss projects the position error of the roll-out at each label instant $j$ onto the expert heading direction, decomposing it into a longitudinal component $e_{x,j}$ and a lateral component $e_{y,j}$, alongside the heading error $e_{\psi,j}$ and the speed error $e_{v,j}$, all defined as the difference between rollouts and demonstration:
\begin{equation}
\begin{aligned}
\mathcal{L}_{\mathrm{state}}=\Big\langle&w_y\operatorname{Hub}_{0.1}(e_{y,j})+w_x\,\ell_\kappa(e_{x,j})\\
&+w_\psi\operatorname{Hub}_{0.02}(e_{\psi,j})+w_v\,\ell_\kappa(e_{v,j})\Big\rangle_\omega+w_R\mathcal{L}_R,
\end{aligned}
\label{eq:state-loss}
\end{equation}
where $\operatorname{Hub}_\varsigma$ is the Huber function with transition point $\varsigma$, $\ell_\kappa$ is the risk-sensitive loss in Eq.~\eqref{eq:asym}, $\langle\cdot\rangle_\omega$ is the mean over samples and valid label instants weighted by the sample weight $\omega$, and $\mathcal{L}_R$ ties each rolled-out position to the lane center via the route label. The sample weight emphasizes  decision-informative rare samples: launch and stop windows, onsets of emergency braking, and hard-steering samples. The control-input anchor
\begin{equation}
\mathcal{L}_{\mathrm{anc}}=\frac{\sum_j\Omega_j\left[\eta_j\operatorname{Hub}_{0.2}\left(u_{x,j}-u^\star_{x,j}\right)+4\left|u_{s,j}-u^\star_{s,j}\right|\right]}{\sum_j\Omega_j}
\label{eq:anchor}
\end{equation}
encodes identifiability: the longitudinal weight $\eta_j$ is strong where the state loss is blind (standstill braking or launch), weak while moving (aliased labels), raised on sustained emergency brakes (clean labels), and zero on recovery samples. The weight $\Omega_j$ combines the sample weight $\omega$ with a decay along the horizon and a raise at standstill. The steering term is L1, robust to aliased steering peaks, and scaled for its smaller range. With weights $w_\Delta$, $w_J$, $w_{a_y}$ and $w_\zeta$, a regularizer $\mathcal{L}_{\mathrm{reg}}$ penalizes knot-to-knot changes, longitudinal jerk, lateral acceleration $a_y$ beyond a comfort limit, and tire utilization $\zeta=\max_i|F_{y,i}|/(\mu_iF_{z,i})$ of Eq.~\eqref{eq:single-track} beyond a limit.

\subsubsection{Closing the Loop}
\label{subsubsec:closed-loop}
Four mechanisms narrow the open-loop-to-closed-loop transfer gap:
\begin{itemize}
    \item \textbf{Risk-sensitive loss.} In braking situations the distribution of future trajectories is bimodal (the vehicle either stops or proceeds), and a symmetric loss converges to the conditional median between the two modes, resulting in braking commands that are delayed. DIVA instead applies an asymmetric quantile loss
    \begin{equation}
    \ell_\kappa(e)=\left(1+\kappa\,\iota\,\mathbb{I}[e>0]\right)|e|,
    \label{eq:asym}
    \end{equation}
    to the longitudinal and speed errors $e_{x,j}$ and $e_{v,j}$, which are positive when the roll-out is ahead of or faster than the expert. The indicator $\iota\in\{0,1\}$ activates the asymmetry ($\kappa=3$) in braking situations; the minimizer is then the $1/(2+\kappa)$ quantile, so the roll-out commits to stopping once the stop-class probability exceeds $1/(2+\kappa)=0.2$. A braking situation is flagged when the vehicle is moving and either (i)~a hazard has been flagged or braked for within the horizon while the speed reference signals a stop, or (ii)~an emergency brake has occurred within 0.5\,s. An auxiliary head predicts the emergency-braking flags with a binary cross-entropy loss $\mathcal{L}_{\mathrm{haz}}$.
    \item \textbf{Causal state feedback.} Including acceleration states in the feedback token introduces a causal shortcut~\cite{wen2020fighting}, since these states effectively encode the previous command and allow the network to copy rather than predict it. This shortcut manifests after each upshift torque interruption, where the closed loop settles into a self-confirming low-acceleration equilibrium. The state-feedback token is therefore restricted to $v_x$ and $r_{\mathrm{avg}}$, where the yaw rate is computed via a backward finite difference to preserve strict causality; a central difference would implicitly leak the next expert state into the current observation.
    \item \textbf{Recovery samples.} With probability 0.3, the initial speed of both the roll-out and $\mathbf{o}$ is perturbed while the targets remain the expert states, so that the controller learns speed feedback from off-trajectory states without additional sensor data.
    \item \textbf{Beyond-preview reference distrust.} With probability 0.5, the reference route beyond the preview distance is shifted along its normal by an offset that grows quadratically from zero at the preview boundary to a random value at the final point, so that the controller learns to rely on its own scene understanding.
\end{itemize}

\subsubsection{Training Under Model Mismatch}
\label{subsubsec:mismatch-training}
In training, $p(\boldsymbol{\theta})$ places probability 0.5 on the nominal value $\boldsymbol{\theta}_0$ and spreads the rest over randomized deviations. The targets remain the nominal expert states, so the controller learns commands that reproduce the expert behavior under mismatch. The controller is informed of the mismatch through the same lumped deviations the observer would report:
\begin{equation}
\mathbf{d}=\nu\left[\bar b_s+n_s,\;\bar d_x+n_x,\;1\right]^\top,\qquad\nu\sim\mathrm{Bernoulli}(0.7),
\label{eq:context}
\end{equation}
where $\bar b_s$ and $\bar d_x$ are the lumped deviations in Eq.~\eqref{eq:lumped}, evaluated once at the steady state of the expert command at $t$ and held over the horizon, $n_s$ and $n_x$ are zero-mean Gaussian noises emulating the estimation error, the last component flags an available estimate, and withholding the estimate ($\nu=0$) keeps the controller valid before observer convergence.

\subsubsection{Gradient Gate for Joint Training}
\label{subsubsec:gradient-gate}
The planner and the control decoder are trained jointly from scratch on the frozen pretrained backbone, and the control loss reaches the planner through the feature interface and the explicit reference in Eq.~\eqref{eq:memory}. On the shared layers, the control-loss gradient is negligible at initialization but grows to dominate the planning-loss gradient at convergence. Every tensor that the control branch receives from the planner therefore passes a gate
\begin{equation}
\mathcal{G}_\lambda(\mathbf{x})=\operatorname{sg}(\mathbf{x})+\lambda\left(\mathbf{x}-\operatorname{sg}(\mathbf{x})\right),
\label{eq:gate}
\end{equation}
where $\operatorname{sg}$ is the stop-gradient operator. The gate is the identity in the forward pass and scales the backward pass by a constant $\lambda=0.1$, which brings the two gradients to the same order at convergence. Unlike a loss weight, it leaves every forward value and the planner's own gradient untouched; $\lambda=0$ reproduces the training signal of a standalone planner exactly. The complete objective is
\begin{equation}
\mathcal{L}=\mathcal{L}_{\mathrm{state}}+w_a\mathcal{L}_{\mathrm{anc}}+\mathcal{L}_{\mathrm{reg}}+w_h\mathcal{L}_{\mathrm{haz}}+w_p\mathcal{L}_{\mathrm{plan}},
\label{eq:total}
\end{equation}
where $\mathcal{L}_{\mathrm{plan}}$ is the route and speed loss of the planning decoder~\cite{transfuser++}.

\subsection{Asynchronous Execution and Online Disturbance Estimation}
\label{subsec:rate}

During training, the controller is evaluated on open-loop roll-outs initialized from recorded states, with perception available at every step. At deployment, however, two additional challenges arise: the control decoder must execute at every control step while perception updates arrive at a lower rate, and the deployed vehicle inevitably deviates from the nominal dynamics model.

\subsubsection{Asynchronous Execution}
\label{subsubsec:full-bandwidth}
Because perception cannot run at the control rate, the tokens in Eq.~\eqref{eq:memory} are split into two groups by update rate. The tokens derived from perception, $\phi_M(\mathbf{M})$, $\phi_P(\mathbf{P})$ and the speed token, are cached at $f_{\mathrm{perc}}$ together with the route and the pose of their sensor frame, while $E_o(\mathbf{o})$, $E_d(\mathbf{d})$ and the route tokens of the vehicle-relative route are recomputed at every control step. Between perception updates, the cached route is transformed to the current pose by a rigid-body transform,
\begin{equation}
\boldsymbol{\rho}'_i=
\begin{bmatrix}
\cos\Delta\psi & \sin\Delta\psi\\
-\sin\Delta\psi & \cos\Delta\psi
\end{bmatrix}
\left(\boldsymbol{\rho}_i-\begin{bmatrix}\Delta X\\\Delta Y\end{bmatrix}\right),
\label{eq:reanchor}
\end{equation}
where $(\Delta X,\Delta Y,\Delta\psi)$ denotes the displacement of the vehicle relative to the sensor frame of the cached route. After arc-length resampling of the transformed route, only the three-layer control decoder needs to be evaluated, constituting the \emph{fast path} of the system. To inform the control decoder of the staleness of its cached perception, a latency embedding encodes both the elapsed time since the last perception update and the accumulated ego-motion $(\Delta X,\Delta Y,\Delta\psi)$ over that interval. This embedding is initialized to zero, so that the decoder's behavior is unaffected when perception is synchronous with control. During training, latency robustness is encouraged by randomly delaying the perception inputs by one or two label periods with probability 0.3. In the limiting case where perception runs at every control step with zero latency, the asynchronous execution path degenerates to standard synchronous operation.

\subsubsection{Online Disturbance Observer}
\label{subsubsec:ukf}
The lumped deviations defined in Eq.~\eqref{eq:lumped} are estimated online by an augmented unscented Kalman filter (UKF). The filter state is $[v_x,v_y,r,b_s,d_x]^\top$, with speed and yaw rate serving as measurements. The process model consists of one control step of $\mathcal{F}$ evaluated at the nominal parameters $\boldsymbol{\theta}_0$, in which $\bar{b}_s$ and $\bar{d}_x$ are augmented as random-walk states; the filter estimates of these augmented states thus converge to $\bar b_s$ and $\bar d_x$. The road grade $\gamma$ is set to zero in the process model so that its effect is absorbed into the longitudinal deviation $d_x$. To prevent uninformative measurements from degrading the estimates, each deviation is frozen in operating regimes where the corresponding measurement channel carries negligible observability, such as low-speed steering for $\bar{b}_s$ or near-standstill braking for $\bar{d}_x$. Additionally, a speed innovation exceeding $T_s$ times the maximum tire-limited acceleration triggers a temporary freeze of both estimates, safeguarding the filter against corruption from collision events. The resulting estimates $\hat b_s$ and $\hat d_x$ are assembled into the disturbance token $\mathbf{d}=[\hat b_s,\hat d_x,1]^\top$ that the control decoder receives as input. Two quantities are deliberately excluded from the estimation: tire-road friction, which is observable from speed and yaw rate only during the brief transients in which the tires saturate and is therefore not reliably estimable in steady driving, and the brake gain deviation, which is not separately identifiable and enters the filter through $\bar d_x$.

\section{Experiments}
\label{sec:experiments}

The experiments address four questions: (1) does extending the differentiable chain to vehicle control improve closed-loop driving and execution quality? (2) which components of DIVA account for the effect? (3) how does the model behave under sustained vehicle dynamics mismatch, compared with classical controllers that use the same identified model and observer? (4) what does the control loop cost, and does it remain effective when perception runs slower than the control rate? Section~\ref{subsec:case-study} presents case studies.

\subsection{Experimental Setup and Baselines}
\label{subsec:setup}

\subsubsection{Benchmark and Training Data}
All closed-loop experiments use the Bench2Drive-220 benchmark~\cite{jia2024bench2drive} with its official metrics: driving score (DS), success rate (SR), efficiency, comfortness and five ability scores. Training follows~\cite{transfuser++} on the PDM-Lite dataset~\cite{sima2024drivelm}, recorded at 4\,Hz, with the same route filter and data split; the perception backbone is pretrained as in~\cite{transfuser++} and frozen.

\subsubsection{Evaluation Subsets}
The ablation, mismatch and perception-rate studies use \emph{Hard-52}, a subset of 52 routes from Bench2Drive-220 that emphasize lateral and longitudinal control complexity: 23 lateral maneuvers, 13 junction turns, 12 merging or highway routes and 4 emergency-braking routes, fixed before any ablation run. On the remaining routes, DIVA and DIVA+PID are nearly tied (92.14 vs.\ 91.20 DS). The model mismatch study additionally uses a traffic-free \emph{maneuver suite} of 16 routes grouped into six families (junction turns, highway, lane changes, stop-and-go, emergency braking, curves). Background traffic and scenario actors are removed except in the emergency-braking family; the evaluation is then nearly deterministic, so tracking metrics resolve the controllers and a single trial per condition suffices.

\subsubsection{Shared Stack and Controllers}
All configurations share the same backbone, rules and safety logic of~\cite{transfuser++}; only the planner and the controller change. Two planners are used: the \emph{baseline planner} (trained alone) and the \emph{joint planner} (trained together with the control decoder of DIVA). The compared configurations are:
\begin{itemize}
    \item \textbf{TF++}~\cite{transfuser++}: the baseline planner with its PID, the most widely used controller of closed-loop driving in Bench2Drive: a lateral PID towards a look-ahead point and a longitudinal PID with full-brake and saturating-throttle branches.
    \item \textbf{MPC+DOB}: the baseline planner with an MPC that shares the identified model, the disturbance observer (DOB) of Section~\ref{subsubsec:ukf}, the parameterization and the horizon of the control decoder, solved at every control step by MPPI~\cite{williams2018information} with 1024 roll-outs; it optimizes a hand-designed tracking cost online instead of applying a learned mapping.
    \item \textbf{DIVA+PID}: the joint planner with the PID of TF++, which isolates the effect of joint training.
    \item \textbf{DOB-PID}: the PID of TF++ extended with the DOB, which subtracts the estimated steering bias and cancels the longitudinal disturbance via the identified map.
    \item \textbf{DIVA w/o observer}: the full DIVA network with the estimate withheld ($\nu=0$).
\end{itemize}
In the mismatch study, all controllers use the joint planner. Control-input behavior cloning is treated as an ablation.

\subsubsection{Metrics and Statistics}
Beyond the official metrics, execution quality is quantified by dynamics metrics (mean jerk, mean $|a_x|$, peak deceleration and lateral acceleration, sideslip angle, comfort-limit exceedances) and actuator-activity and tracking metrics (wide-open-throttle (WOT) fraction, throttle variation, brake switches, steering rate, speed error, lateral offset), all macro-averaged over routes. Variants are compared by paired per-route differences with a 95\% CI.

\subsection{Closed-Loop Results on Bench2Drive-220}
\label{subsec:main-results}

\begin{table*}[!t]
\caption{Closed-Loop Driving and Execution Results on Bench2Drive-220}
\label{tab:main}
\centering
\scriptsize
\setlength{\tabcolsep}{2.7pt}
\newcommand{\hd}[1]{\begin{tabular}[b]{@{}c@{}}#1\end{tabular}}
\begin{tabular}{@{}llccccccccccccccccc@{}}
\toprule
 & & \multicolumn{4}{c}{Official metrics} & \multicolumn{5}{c}{Ability (\%) $\uparrow$} & \multicolumn{7}{c}{Execution metrics $\downarrow$} & \\
\cmidrule(lr){3-6}\cmidrule(lr){7-11}\cmidrule(lr){12-18}
Method & Planner & \hd{DS\\$\uparrow$} & \hd{SR\\(\%) $\uparrow$} & \hd{Effi.\\$\uparrow$} & \hd{Comf.\\$\uparrow$} & \hd{Over-\\taking} & \hd{Merg-\\ing} & \hd{Emerg.\\brake} & \hd{Give\\way} & \hd{Traffic\\signs} & \hd{Jerk\\(m/s$^3$)} & \hd{$|a_x|$\\(m/s$^2$)} & \hd{WOT\\frac.} & \hd{Brake sw.\\(1/min)} & \hd{Steering\\rate (1/s)} & \hd{Lat.\ off.\\p95 (m)} & \hd{Speed err.\\(m/s)} & \hd{Speed\\(m/s)}\\
\midrule
TF++~\cite{transfuser++} & baseline & 83.34 & 67.6 & 250.97 & 33.9 & 55.56 & 58.75 & 78.33 & \textbf{50.00} & 86.84 & 16.42 & 4.54 & 0.823 & 42.7 & 0.173 & 0.159 & 3.36 & 8.67\\
MPC+DOB & baseline & 79.59 & 61.2 & \textbf{259.22} & \textbf{38.5} & 37.78 & 53.75 & 76.67 & \textbf{50.00} & 82.63 & 10.92 & 3.74 & \textbf{0.113} & 15.8 & \textbf{0.097} & 0.220 & 3.33 & 8.67\\
DIVA+PID & joint & 83.80 & 67.1 & 248.31 & 31.4 & \textbf{60.00} & 56.25 & 78.33 & \textbf{50.00} & 82.63 & 16.72 & 4.57 & 0.821 & 42.8 & 0.169 & 0.126 & 3.04 & 8.72\\
DIVA (ours) & joint & \textbf{86.73} & \textbf{71.7} & 256.29 & 36.6 & 55.56 & \textbf{65.82} & \textbf{90.00} & 40.00 & \textbf{87.89} & \textbf{8.86} & \textbf{3.19} & 0.203 & \textbf{15.4} & 0.098 & \textbf{0.100} & \textbf{2.96} & 8.63\\
\bottomrule
\end{tabular}
\begin{flushleft}
\scriptsize Effi.: efficiency; Comf.: comfortness; Jerk, $|a_x|$: means; WOT: wide-open throttle; Brake sw.: brake switches; Lat.\ off.: lateral offset; Speed err.: speed error; Speed: mean driving speed (not ranked). Execution metrics are macro-averaged over routes. Baseline planner: trained alone; joint planner: trained with the control decoder of DIVA. 
\end{flushleft}
\end{table*}

In Table~\ref{tab:main}, TF++ and DIVA+PID show that joint training neither improves nor degrades the planner for a hand-designed tracker, so the four configurations differ only in the controller. DIVA attains the highest DS (86.73) and SR (71.7\%), with the largest gains on emergency braking and merging. It also records fewer vehicle collisions and timeouts than DIVA+PID. MPC+DOB obtains the best comfortness and efficiency but loses substantially in DS, with failures concentrated in overtaking scenarios. Because MPC+DOB tracks the full planned route over a 2\,s horizon, it is sensitive to the noisy beyond-preview reference that the learned planner provides. The PID of TF++ retains its score through short preview and saturated commands; DIVA, trained on the planner's actual output distribution, achieves both task performance and smoothness.

\begin{figure}[!t]
\centering
\includegraphics[width=\columnwidth]{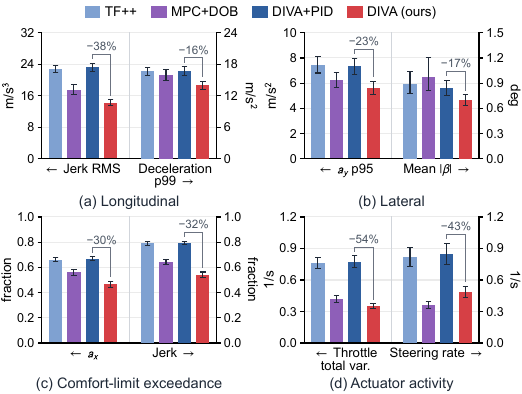}
\caption{Execution metrics on Bench2Drive-220: (a) jerk RMS and deceleration p99, (b) lateral acceleration and mean sideslip angle, (c) comfort-limit exceedance fraction for $a_x$ and jerk, (d) throttle total variation and steering-rate. TF++ and MPC+DOB: baseline planner; DIVA+PID and DIVA: joint planner. }
\label{fig:exec}
\end{figure}

In execution quality (Table~\ref{tab:main}, Fig.~\ref{fig:exec}), against DIVA+PID, which shares its planner, DIVA reduces the mean jerk by 47\%, the brake switches by 64\%, the lateral offset by 21\% and the steering rate by 42\%, while maintaining comparable speed and tracking accuracy. The improvement persists on collision-free routes, confirming that it does not stem from fewer impacts. MPC+DOB matches DIVA in throttle and steering smoothness but tracks the path worst of all configurations. In the secondary metrics of Fig.~\ref{fig:exec}, DIVA is lowest in seven of the eight indicators.

\subsection{Ablation Study}
\label{subsec:ablation}

Each variant repeats training with one change, except where noted, and is evaluated on Hard-52 on its own jointly trained planner.
\begin{itemize}
    \item \textbf{Vehicle model} (Section~\ref{subsec:dynamics}): \emph{w/o vehicle model, control-input BC} (loss on the expert's 4\,Hz commands; recovery samples and randomized dynamics, which require the model, disabled), \emph{nominal kinematic model} (the constant-parameter kinematic bicycle of~\cite{chen2021learning}, evaluated without observer since no identified model exists), \emph{w/o on-policy re-identification} (the calibrated model) and \emph{w/o observer} ($\nu=0$, Section~\ref{subsubsec:mismatch-training}).
    \item \textbf{Gradient into the planner} (Section~\ref{subsubsec:gradient-gate}): $\lambda=0$ and $\lambda=1$.
    \item \textbf{Training strategy} (Section~\ref{subsubsec:closed-loop}): \emph{w/o risk-sensitive loss} ($\kappa=0$), \emph{non-causal state-feedback input} (central differences and acceleration inputs restored), \emph{w/o recovery samples}, and \emph{w/o beyond-preview reference noise}, which also removes the emphasis on hard-steering samples.
    \item \textbf{Temporal discretization} (Section~\ref{subsubsec:three-clocks}): \emph{ZOH, 0.25\,s step}, i.e., zero-order hold of the knots together with one integration step per knot.
\end{itemize}
Table~\ref{tab:ablation} reports the DS with its paired difference to DIVA, closed-loop execution metrics, and offline metrics computed through a common reference roll-out.

\begin{table*}[!t]
\caption{Ablation on Hard-52: Driving Score, Closed-Loop and Offline Metrics}
\label{tab:ablation}
\centering
\footnotesize
\setlength{\tabcolsep}{2.6pt}
\begin{tabular}{@{}lcccccccccccc@{}}
\toprule
 & \multicolumn{9}{c}{Closed loop} & \multicolumn{3}{c}{Offline, 2\,s horizon}\\
\cmidrule(lr){2-10}\cmidrule(l){11-13}
 & & & & & & WOT & Lat. & Brake & Speed & Lat. & & $u_{s,0}$\\
 & & $\Delta$DS $\uparrow$ & SR $\uparrow$ & & Jerk $\downarrow$ & frac.\ & offset $\downarrow$ & sw.\ $\downarrow$ & err.\ $\downarrow$ & p90 $\downarrow$ & Speed $\downarrow$ & error\\
Variant & DS $\uparrow$ & [95\% CI] & (\%) & RC $\uparrow$ & (m/s$^3$) & $\downarrow$ & p95 (m) & (1/min) & (m/s) & (m) & (m/s) & $\downarrow$\\
\midrule
DIVA (ours) & 69.3 & -- & \textbf{36.5} & 91.69 & 10.84 & 0.21 & 0.109 & 17.8 & 4.65 & \textbf{0.306} & 1.110 & 0.0145\\
DIVA+PID & 60.1 & $-9.27$ [$-16.6$, $-2.0$]$^{\dagger}$ & 17.3 & 90.78 & 15.87 & 0.82 & 0.143 & 33.5 & 4.64 & -- & -- & --\\
\midrule
w/o vehicle model, control-input BC & 62.0 & $-7.36$ [$-14.3$, $-0.5$]$^{\dagger}$ & 21.2 & 91.78 & 12.78 & 0.11 & 0.159 & 24.2 & 4.54 & 0.543 & 1.345 & \textbf{0.0094}\\
nominal kinematic model & 68.6 & $-0.77$ [$-8.1$, $+6.6$] & 30.8 & 91.07 & 16.24 & 0.59 & 0.147 & 24.3 & 4.49 & 0.783 & 3.283 & 0.0145\\
w/o on-policy re-identification & 66.3 & $-3.02$ [$-11.7$, $+6.3$] & 28.8 & 91.27 & \textbf{9.71} & 0.21 & 0.127 & 14.3 & 4.61 & 0.351 & 1.153 & 0.0142\\
w/o observer & \textbf{70.0} & $+0.63$ [$-4.0$, $+5.0$] & \textbf{36.5} & 91.47 & 10.86 & 0.27 & 0.112 & 18.6 & 4.43 & -- & -- & --\\
\midrule
$\lambda=0$, planner stop-gradient & 65.5 & $-3.80$ [$-11.3$, $+3.7$] & 28.8 & \textbf{92.45} & 10.32 & 0.23 & 0.174 & 17.0 & 4.56 & 0.365 & 1.155 & 0.0148\\
$\lambda=1$, full gradient & 60.7 & $-8.68$ [$-15.9$, $-2.0$]$^{\dagger}$ & 19.2 & 89.70 & 11.43 & 0.22 & 0.188 & 17.4 & 4.55 & 0.363 & 1.106 & 0.0153\\
\midrule
w/o risk-sensitive loss & 65.0 & $-4.29$ [$-12.4$, $+3.9$] & 26.9 & 91.77 & 11.44 & 0.24 & \textbf{0.104} & 15.0 & 4.61 & 0.312 & \textbf{1.003} & 0.0121\\
non-causal state-feedback input & 60.3 & $-9.04$ [$-19.1$, $+0.9$] & 23.1 & 81.90 & 10.91 & 0.06 & 0.154 & \textbf{13.6} & 5.74 & 0.316 & 1.071 & 0.0136\\
w/o recovery samples & 64.7 & $-4.65$ [$-10.7$, $+1.2$] & 26.9 & 91.89 & 11.24 & 0.22 & 0.156 & 16.8 & 4.77 & 0.340 & 1.097 & 0.0149\\
w/o beyond-preview reference noise & 65.2 & $-4.16$ [$-11.2$, $+3.1$] & 25.0 & 90.95 & 10.46 & 0.21 & 0.156 & 16.6 & \textbf{4.18} & 0.338 & 1.104 & 0.0166\\
\midrule
ZOH, 0.25\,s step & 48.5 & $-20.98$ [$-31.9$, $-10.1$]$^{\dagger}$ & 19.6 & 65.21 & 11.31 & \textbf{0.00} & 0.170 & 22.7 & 6.75 & 1.536 & 1.570 & 0.0299\\
\bottomrule
\end{tabular}
\begin{flushleft}
\footnotesize $\Delta$DS: paired difference to DIVA with bootstrap 95\% CI; $^{\dagger}$: CI excludes zero. Every variant runs on its own jointly trained planner; DIVA+PID uses the planner of DIVA. Offline metrics: 2\,s lateral error (p90), 2\,s speed error (mean) and steering-knot error $u_{s,0}$ (mean absolute) on 6000 validation frames.
\end{flushleft}
\end{table*}

Four variants incur a statistically significant DS loss (bootstrap 95\% CI excludes zero): DIVA+PID, control-input behavior cloning, $\lambda=1$, and the coarse-rate scheme. The ablation yields five findings. First, control-input behavior cloning achieves the lowest offline control-input error yet suffers a significant DS loss, confirming that minimizing the control-input residual does not guarantee closed-loop performance. Second, replacing the identified model with a nominal kinematic bicycle preserves DS but degrades execution quality to the level of DIVA+PID, while omitting on-policy re-identification increases static collisions, indicating that execution fidelity demands an accurate, distribution-matched vehicle model. Third, removing the state-loss gradient entirely ($\lambda=0$) costs DS only within noise, whereas the full gradient ($\lambda=1$) causes significant DS loss and the largest lateral offset, as the unattenuated control loss distorts the planner's route (mean point error 0.207 vs.\ 0.165\,m for $\lambda\le0.1$). Fourth, the non-causal state feedback, the risk-sensitive loss, and the recovery samples are each indistinguishable from DIVA offline, yet their removal consistently degrades closed-loop DS; notably, the coarse-rate scheme suffers the largest DS loss of all variants through route-completion collapse and tripled timeouts rather than collisions. Fifth, the observer adds no measurable benefit at nominal conditions; its value emerges under strong mismatch.

\subsection{Robustness to Vehicle Dynamics Model Mismatch}
\label{subsec:mismatch}

\begin{figure*}[!t]
\centering
\includegraphics[width=\textwidth]{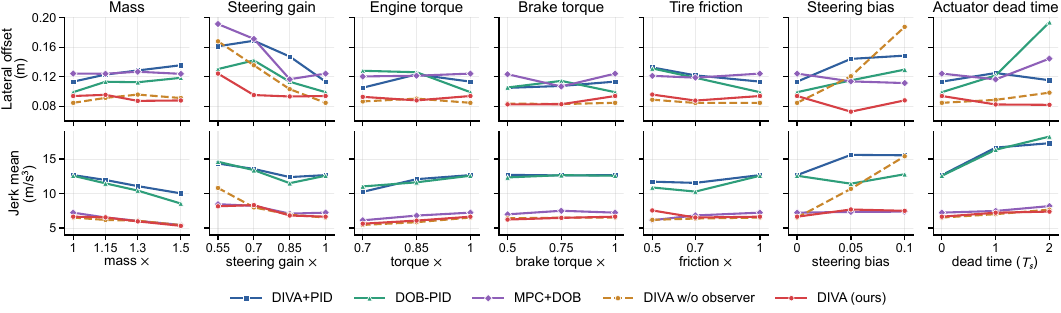}
\caption{Lateral offset (p95, top) and mean jerk (bottom) under sustained vehicle-side mismatch on the traffic-free maneuver suite (16 routes). Columns: perturbed quantity with nominal at factor 1, bias 0 or dead time 0. All controllers on the joint planner.}
\label{fig:mismatch}
\end{figure*}

\begin{figure*}[!t]
\centering
\includegraphics[width=\textwidth]{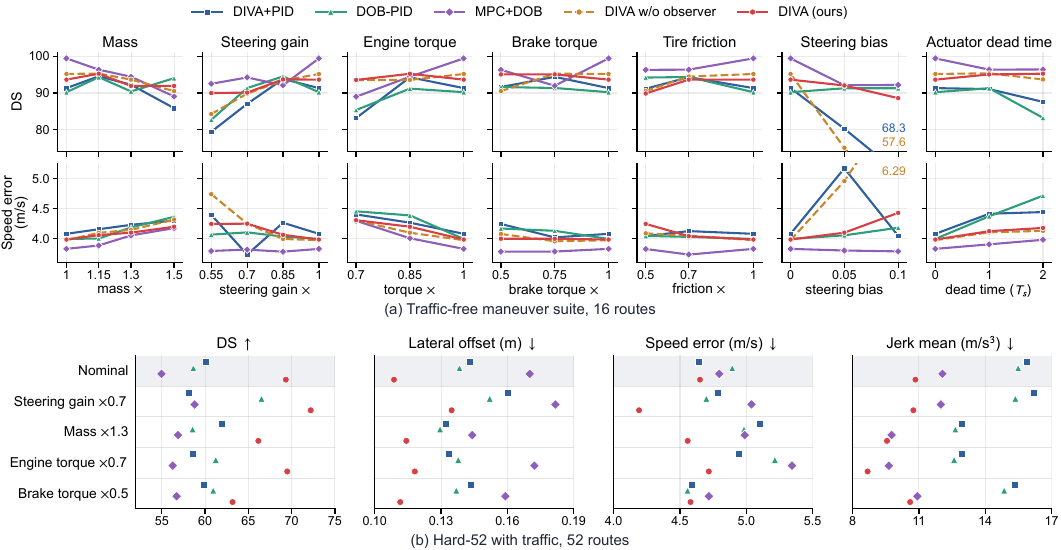}
\caption{DS and tracking under vehicle-side mismatch, all controllers on the joint planner. (a) Maneuver suite: DS (top) and speed error (bottom) vs.\ perturbed quantity. (b) Hard-52 with traffic: DS, lateral offset p95, speed error and jerk for the nominal vehicle (shaded) and one strength of four perturbation families.}
\label{fig:mismatch-ds}
\end{figure*}

The mass, steering gain, engine torque, brake torque and tire friction of the deployed vehicle are scaled at initialization, and a steering bias and an actuator dead time are injected between the controller and the vehicle. Two tiers are used: the traffic-free maneuver suite with 16 perturbations spanning seven families (Figs.~\ref{fig:mismatch}--\ref{fig:mismatch-ds}(a)), and Hard-52 with traffic under one strength of four families (Fig.~\ref{fig:mismatch-ds}(b)). On the suite, DIVA achieves the lowest lateral offset in all 17 conditions and the lowest jerk in most (Fig.~\ref{fig:mismatch}). On Hard-52, where interaction with traffic dominates the score, DIVA does not collapse: it attains the highest DS and the lowest lateral offset in every perturbed condition. 

\begin{figure}[!t]
\centering
\includegraphics[width=\columnwidth]{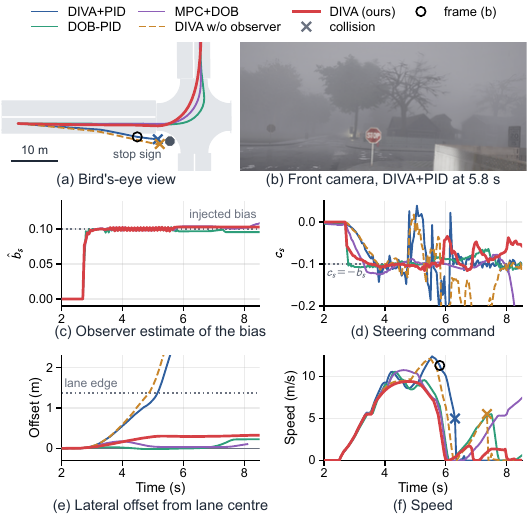}
\caption{Steering bias $+0.10$ on suite route 2084 (junction left turn at night). (a) Paths of the five controllers. (b) Front camera of DIVA+PID at 5.8\,s, 0.5\,s before its collision. (c) Observer estimate $\hat b_s$ against the injected bias. (d) Steering command; dotted: exact compensation $c_s=-b_s$. (e) Lateral offset from the lane center; dotted: lane edge. (f) Speed. DIVA+PID and DIVA w/o observer use no estimate.}
\label{fig:mismatch-case}
\end{figure}

\begin{figure}[!t]
\centering
\includegraphics[width=\columnwidth]{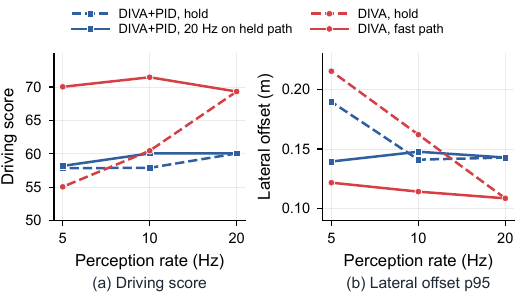}
\caption{DS (a) and lateral offset p95 (b) on Hard-52 at reduced perception rates with 0.1\,s latency; at 20\,Hz, the synchronous runs without latency, where hold and fast path coincide. Hold: command held at the perception rate; fast path: asynchronous DIVA; DIVA+PID on held path: its PID at 20\,Hz on the last perception result transformed to the current pose.}
\label{fig:perception-rate}
\end{figure}

\begin{figure*}[!t]
\centering
\includegraphics[width=\textwidth]{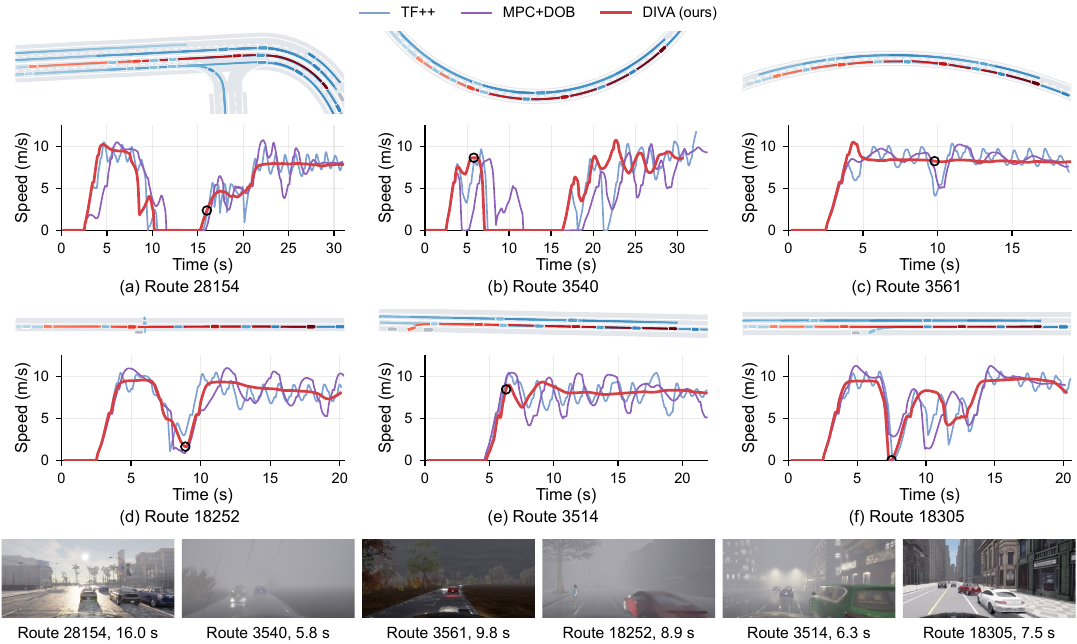}
\caption{Case studies on six routes where all controllers score 100; TF++ and MPC+DOB run on the baseline planner. (a) T-junction stop and turn. (b) Hard-braking lead vehicle. (c) Control-loss disturbance. (d) Pedestrian stepping out. (e) Parking exit into traffic. (f) Vehicle cut-in. Bottom row: front camera of DIVA on each route.}
\label{fig:case-study}
\end{figure*}

Under reduced steering gain and steering bias, the two controllers receiving the observer's estimate (DOB-PID and DIVA) degrade least, whereas DIVA+PID and DIVA w/o observer suffer large lateral-offset increases and substantial DS losses, confirming that the observer determines where the mismatch is observable. Fig.~\ref{fig:mismatch-case} illustrates a representative case: the observer converges within 0.25\,s and the observer-equipped controllers compensate the bias immediately, while the others depart the lane and collide with roadside infrastructure. DOB-PID achieves comparable path-holding but retains the elevated jerk of DIVA+PID. Under actuator dead time, which lies outside the training distribution $p(\boldsymbol{\theta})$, DOB-PID degrades sharply in both lateral offset and jerk, whereas DIVA remains stable, indicating that the amortized receding-horizon controller is inherently more tolerant of unmodeled dead time than direct feedforward compensation. With the clean reference of the traffic-free suite, MPC+DOB attains the highest DS in most conditions, but with the noisy reference of a learned planner in traffic it yields the lowest (Table~\ref{tab:main}), revealing that its advantage lies in tracking rather than reference robustness. Finally, friction constitutes a negative result: it is not observable over the brief transients in which the tires saturate, the observer carries no channel for it, and the disturbance estimate therefore cannot compensate for friction variations.

\subsection{Latency and Perception-Rate Decoupling}
\label{subsec:latency}

Table~\ref{tab:latency} shows that perception dominates every cycle at about 16\,ms, while the fast path of DIVA costs less than 0.5\,ms; all configurations except MPC+DOB with the GPU solver meet the 50\,ms budget of 20\,Hz. Fig.~\ref{fig:perception-rate} evaluates the closed loop at reduced perception rates: with the fast path the DS of DIVA is insensitive to the perception rate and its lateral offset grows only slowly, whereas the same network with its command zero-order held collapses at 5\,Hz. DIVA+PID runs at 20\,Hz on the held path, transformed to the current pose, with the current speed. It stays close to its synchronous DS.

\subsection{Case Studies}
\label{subsec:case-study}

\begin{table}[!t]
\caption{Latency of the Complete Control Cycle}
\label{tab:latency}
\centering
\footnotesize
\setlength{\tabcolsep}{4pt}
\begin{tabular}{@{}lccc@{}}
\toprule
Configuration & Mean (ms) & p95 (ms) & Rate (Hz)\\
\midrule
DIVA+PID & 17.28 & 17.97 & 58\\
DOB-PID & 18.08 & 20.60 & 55\\
MPC+DOB, CPU & 45.08 & 45.92 & 22\\
MPC+DOB, GPU & 81.47 & 83.79 & 12\\
DIVA, synchronous & 21.30 & 23.93 & 47\\
DIVA, async., 20\,Hz perception & 18.44 & 19.19 & 54\\
DIVA, 10\,Hz perc.: perception cycle & 18.57 & 19.25 & 54\\
DIVA, 10\,Hz perc.: control-only cycle & \textbf{3.78} & \textbf{4.09} & \textbf{265}\\
Fast path without CUDA graph & 4.78 & 5.22 & 209\\
\bottomrule
\end{tabular}
\begin{flushleft}
\footnotesize RTX 5090, Core Ultra 9 285K, PyTorch 2.7.1, CUDA 12.8, 32-bit. The control-only cycle of DIVA runs only the fast path.
\end{flushleft}
\end{table}

On the six routes of Fig.~\ref{fig:case-study}, all controllers score 100 and finish in similar time, so that the execution differences are comparable. DIVA attains the lowest jerk on every route and far fewer brake switches than TF++, which alternates relay-like between wide-open throttle and full brake. Behind the hard-braking lead vehicle, TF++ stops and relaunches repeatedly, DIVA only once; under the control-loss disturbance, DIVA holds its cruise speed, whereas TF++ and MPC+DOB suffer large speed drops; after the pedestrian and parking-exit encounters, DIVA settles to a steadier speed than both.

\section{Conclusion}
\label{sec:discussion}

DIVA bridges the planning-to-control gap in end-to-end autonomous driving by jointly training the controller through identified differentiable vehicle dynamics. Supervising on states rather than commands decouples the demonstrations from the recording vehicle's actuators: a dedicated controller can be trained for any identified vehicle on the same demonstration set in a single run, without per-vehicle controller-tuning. The identified model and observer are independently valuable, already conferring mismatch robustness on classical controllers, while the learned controller adds smooth, actuator-aware execution at the control rate with a sub-millisecond control-only cycle decoupled from perception latency. On Bench2Drive, DIVA achieves a higher driving score than the most widely adopted baseline with substantially smoother execution. The ablations confirm that task performance requires a vehicle dynamics model, execution quality depends on its fidelity, and closed-loop viability demands rate consistency across control-input parameterization, supervision, and execution. Future work will transfer the framework to real vehicles, add observer channels for friction and brake effectiveness, and extend this method into wider end-to-end paradigms.

\appendix

\subsection{Vehicle Dynamics Model and Identification}
\label{app:parameters}
This appendix gives the complete model of Section~\ref{subsec:dynamics}, its identification and its parameters.
\begin{figure}[!t]
\centering
\includegraphics[width=\columnwidth]{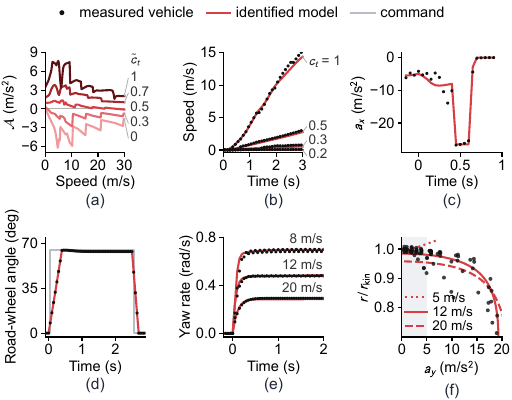}
\caption{Identification experiments: measured vehicle against model roll-outs under the same commands. (a) Acceleration map; (b) launch from rest; (c) full-brake step; (d) steering slew-rate limiter; (e) yaw-rate steps up to a lateral acceleration $a_y\approx6$\,m/s$^2$; (f) steady-state yaw-rate gain in all 86 step-steer tests (dots) and of the model at 5, 12 and 20\,m/s (lines; shaded: benchmark range).}
\label{fig:identification}
\end{figure}

\subsubsection{Model Equations}
Each actuator command $c\in\{c_s,c_t,c_b\}$ is processed by a slew-rate limiter that advances its rate-limited counterpart $\tilde c$ towards $c$ by at most $\dot c^{\uparrow}T_s$ per step when the input magnitude increases and $\dot c^{\downarrow}T_s$ when it decreases, where $\dot c^{\uparrow}$ and $\dot c^{\downarrow}$ denote the apply and release rates, respectively. The powertrain and brake accelerations $a_d$ and $a_b$ track their respective references $\mathcal{A}(v_x,\tilde c_t)$ and $a_{\mathrm{brk}}\tilde c_b$ through exactly discretized first-order lags with time constants $\tau_d$ and $\tau_b$. The map $\mathcal{A}$ is bilinearly interpolated; although the underlying shift schedule exhibits hysteresis, omitting the gear state does not degrade the open-loop accuracy.
\begin{table}[!t]  %
\caption{Parameters of the Identified Vehicle Model}
\label{tab:parameters}
\centering
\footnotesize
\setlength{\tabcolsep}{3pt}
\begin{tabularx}{\columnwidth}{@{}l>{\raggedright\arraybackslash}Xl@{}}
\toprule
Symbol & Quantity & Value\\
\midrule
\multicolumn{3}{@{}l}{\emph{Specification}}\\
$m$ & Mass & 1696\,kg\\
$l_f$, $l_r$ & Center of mass to front, rear axle & 1.172, 1.689\,m\\
$W$ & Track width & 1.593\,m\\
$x_o$ & Center of mass ahead of reference point & 0.30\,m\\
$\delta_{\max}(v_x)$ & At 0, 20, 60, 120\,km/h & 70, 63, 56, 49$^\circ$\\
\midrule
\multicolumn{3}{@{}l}{\emph{Calibrated}}\\
$\dot c^{\uparrow}$, $\dot c^{\downarrow}$ & Apply, release rate of steering & 2.5, 5.0\,s$^{-1}$\\
 & \quad of throttle and brake & 6, 10\,s$^{-1}$\\
$\tau_d$, $\tau_b$ & Powertrain, brake lag: calibrated & 0.15, 0.15\,s\\
 & \quad re-identified & 0.10, 0.05\,s\\
$a_{\mathrm{brk}}$ & Deceleration per unit brake input & $-4.83$\,m/s$^2$\\
$a_{\mathrm{lock}}$ & Locked-wheel deceleration & $-26.5$\,m/s$^2$\\
$v_{\mathrm{g1}}$ & Upper speed of first gear under the shift schedule & 5.5\,m/s\\
$[v_{\mathrm{lo}},v_{\mathrm{hi}}]$ & Non-locking speed band in first gear & [1.8, 4.6]\,m/s\\
$\mathcal{A}$ & Grid: speed (0--40\,m/s) $\times$ throttle & 161$\times$21\\
\multicolumn{3}{@{}l@{}}{\begin{tabular}{@{}lccccc@{}}
Upshift $q\to q{+}1$ & 1 & 2 & 3 & 4 & 5\\
$v^{\mathrm{up}}_q(\tilde c_t)$ (m/s) & 5.1--6.0 & 8.4--11.3 & 13.5--17.9 & 18.0--22.5 & 26.2--32.6\\
$t^{\mathrm{sh}}_q$ (s) & 0.122 & 0.017 & 0.058 & 0.024 & 0.077\\
$\sigma_q$ (m/s) & 0.34 & 1.0 & 1.0 & 0.63 & 0.25\\
\end{tabular}}\\
\midrule
\multicolumn{3}{@{}l}{\emph{Identified}}\\
$I_z$ & Yaw moment of inertia & 2377\,kg\,m$^2$\\
$C_f$, $C_r$ & Cornering stiffness & 125.3, 90.5\,kN/rad\\
$\mu_f$, $\mu_r$ & Lateral force saturation coefficient & 2.40, 2.65\\
\bottomrule
\end{tabularx}
\end{table}

Two discrete regimes supplement the longitudinal channel. A wheel-lock regime ($\chi=1$) engages when the brake input exceeds 0.02 in the first gear at speeds between 0.05\,m/s and the maximum first-gear speed $v_{\mathrm{g1}}$, outside the non-locking band $[v_{\mathrm{lo}},v_{\mathrm{hi}}]$; while active, it overrides $a_x$ with the locked-wheel deceleration $a_{\mathrm{lock}}$ and persists until the brake is released or the vehicle stops. A standstill-hold regime constrains $v_x\ge0$ and resets $a_d$ to zero when $v_x<0.02$\,m/s and the brake input exceeds 0.005.

In the lateral channel, the inner road-wheel steer angle is $\delta_{\mathrm{in}}=\delta_{\max}(v_x)\,\tilde c_s$. The equivalent single-track angle $\delta$ follows from the Ackermann relation $\cot\delta=\cot\delta_{\mathrm{in}}+W/(2L)$, with track width $W$ and wheelbase $L=l_f+l_r$, and is computed as $\delta=\arctan\big(\tan\delta_{\mathrm{in}}/(1+|\tan\delta_{\mathrm{in}}|W/(2L))\big)$, an odd, continuously differentiable approximation. The single-track model is
\begin{gather}
\alpha_f=\delta-\arctan\frac{v_y+l_fr}{v_x^+},\qquad
\alpha_r=-\arctan\frac{v_y-l_rr}{v_x^+},\notag\\
F_{y,i}=\mu_iF_{z,i}\tanh\frac{C_i\alpha_i}{\mu_iF_{z,i}},\qquad i\in\{f,r\},
\label{eq:single-track}\\
\dot v_y=\frac{F_{y,f}\cos\delta+F_{y,r}}{m}-v_xr,\quad
\dot r=\frac{l_fF_{y,f}\cos\delta-l_rF_{y,r}}{I_z},\notag
\end{gather}
where $v_x^+=\max(v_x,1\,\mathrm{m/s})$, $\alpha_i$ are the axle slip angles, $F_{y,i}$ the lateral axle forces, and $F_{z,f}=mgl_r/L$ and $F_{z,r}=mgl_f/L$ the static axle loads; Table~\ref{tab:parameters} lists the other symbols. At low speed, $(v_y,r)$ blends into the kinematic values $v_{y,\mathrm{kin}}=v_x\tan\beta_{\mathrm{kin}}$ and $r_{\mathrm{kin}}=v_x\tan\delta\cos\beta_{\mathrm{kin}}/L$, with $\beta_{\mathrm{kin}}=\arctan(l_r\tan\delta/L)$. The blending weight $\varpi=\min\{\max\{(v_x-1)/2,0\},1\}$ moves from purely kinematic behavior below 1\,m/s to purely dynamic behavior above 3\,m/s. The pose follows $\dot X=v_x\cos\psi-(v_y-x_or)\sin\psi$, $\dot Y=v_x\sin\psi+(v_y-x_or)\cos\psi$ and $\dot\psi=r$. The lateral and pose channels advance by $N_s=5$ semi-implicit Euler sub-steps per control step.

\subsubsection{Identification Protocol}
\emph{Specification} parameters, namely the mass, the position of the center of mass, the wheelbase, the track width, $\delta_{\max}(v_x)$ and the gear ratios, are taken from the vehicle specification. \emph{Calibrated} parameters are measured with constant-input, step and coast-down experiments on a flat road: the map $\mathcal{A}$ (Fig.~\ref{fig:identification}(a)), $a_{\mathrm{brk}}$, the slew-rate limits (Fig.~\ref{fig:identification}(c), (d)), the lags, the dead bands (Fig.~\ref{fig:identification}(b)), the shift schedule, the lock regime (Fig.~\ref{fig:identification}(c)) and the standstill behavior. \emph{Identified} parameters, which no single experiment exposes, are estimated from step-steer tests (Fig.~\ref{fig:identification}(e), (f)) and swept-sine steering excitations: $I_z$, $C_f$, $C_r$, $\mu_f$ and $\mu_r$. Dead times, brake deceleration, shift speeds and yaw-rate gains re-measured on a second test platform agree with the fitted values to three or four significant digits. Drives with random excitation are withheld from every fit for held-out validation.

\subsubsection{On-Policy Re-Identification}
Under the first controller's commands, the calibrated model reveals four error sources (Fig.~\ref{fig:onpolicy}(b)--(d)): prolonged upshift torque interruptions, map artifacts that cap output at moderate throttle, fixed rather than throttle-dependent shift points, and an overestimated brake-release lag. To address these, for $\tilde c_t\ge0.5$ the cells below the first upshift are raised to the in-gear value and the gear boundaries are aligned with the measured upshift speeds $v^{\mathrm{up}}_q(\tilde c_t)$. Where $\mathcal{A}>0$, the expected time lost per upshift is folded into the map via
\begin{equation}
\mathcal{A}'\!\left(v_x,\tilde c_t\right)=\frac{\mathcal{A}}{1+\mathcal{A}\sum_{q=1}^{5}t^{\mathrm{sh}}_q\,\mathcal{N}\!\left(v_x-v^{\mathrm{up}}_q(\tilde c_t);0,\sigma_q^2\right)},
\label{eq:shift-pause}
\end{equation}
where $\mathcal{A}$ abbreviates $\mathcal{A}(v_x,\tilde c_t)$, $t^{\mathrm{sh}}_q$ is the net upshift delay from gear $q$, and $\mathcal{N}(\cdot;0,\sigma_q^2)$ is a Gaussian kernel of width $\sigma_q$; the expression adds the expected pause to the reciprocal acceleration $1/\mathcal{A}$. The lag constants $(\tau_d,\tau_b)$ are selected by grid search on the open-loop speed error over training routes, while the lateral model, brake gain, and lock regime remain unchanged. The held-out drives were used only for structure selection; thus Fig.~\ref{fig:onpolicy}(f) constitutes a validation rather than a test result.

\subsubsection{Parameters}
Table~\ref{tab:parameters} lists the parameters of $\mathcal{F}$ for the vehicle of Bench2Drive in CARLA 0.9.15 at $T_s=0.05$\,s, grouped by tier. About 7\% of the command changes take effect one control period late, presumably because the client applies them asynchronously to the simulation step. For the calibrated model, the 1\,s position error on the held-out drives concentrates in discrete longitudinal events: low-speed braking and lock-up (0.585\,m), gear shifts (0.411\,m) and launches (0.257\,m), ahead of high lateral acceleration (0.222\,m), grades (0.180\,m) and late commands (0.151\,m); event-free windows show 0.057\,m.

\subsection{Implementation Details and Hyperparameters}
\label{app:implementation}
Table~\ref{tab:hyper} lists the settings of DIVA, shared by all ablation variants except for their defining change. The yaw rate $r_{\mathrm{avg}}$ in $\mathbf{o}$ is a backward heading difference over $T_{\mathrm{lab}}$ in training and the mean measured yaw rate over the last five control steps at deployment.

\begin{table}[!t]
\caption{Implementation Details and Hyperparameters}
\label{tab:hyper}
\centering
\footnotesize
\setlength{\tabcolsep}{3pt}
\begin{tabularx}{\columnwidth}{@{}>{\raggedright\arraybackslash}Xl@{}}
\toprule
Item & Value\\
\midrule
Token width, embedding MLP width & 256, 128\\
Planning decoder: heads, FFN, GRU width & 8, 2048, 64\\
Control decoder: heads, FFN, dropout & 8, 1024, 0.1\\
$w_y$, $w_x$, $w_\psi$, $w_v$, $w_R$ & 1.0, 0.5, 2.0, 0.5, 0.5\\
$w_a$, $w_h$, $w_p$ & 0.5, 0.2, 1.0\\
Anchor $\Omega_j$: decay per knot, factor at standstill & 0.8, 4\\
Anchor $\eta_j$: standstill or launch, moving, emergency brake & 1, 0.2, 0.6\\
$w_\Delta$, $w_J$, $w_{a_y}$, $w_\zeta$ & 0.05, $2\times10^{-3}$, 0.01, 1.0\\
Positive-class weight in $\mathcal{L}_{\mathrm{haz}}$ & 3\\
Randomized $\ln k_s$, $\ln k_d$, $\ln k_b$, $\ln k_m$ & $\pm0.25$\\
Randomized $b_s$, $b_t$, $k_\mu$, $d_x$ & $\pm0.08$, $\pm0.05$, [0.4, 1], $\pm1$\,m/s$^2$\\
Noise s.d.\ of $n_s$, $n_x$ & 0.01, 0.15\,m/s$^2$\\
Relative noise on $\mathbf{o}$ & 5\%\\
Learning rate, warm-up steps, cosine floor & $3\times10^{-4}$, 1000, 1\%\\
Gradient clipping: control decoder, planner & 1.0, 5000\\
Batch size, training steps & 32$\times$8 GPUs, 58\,380\\
UKF process s.d.: $v_x$, $v_y$, $r$, $b_s$, $d_x$ & 0.05, 0.05, 0.02, 0.002, 0.03\\
UKF measurement s.d.: $v_x$, $r$ & 0.05, 0.01\\
UKF initial s.d.: $v_x$, $v_y$, $r$, $b_s$, $d_x$ & 0.32, 0.32, 0.22, 0.05, 0.5\\
UKF limits of $|\hat b_s|$, $|\hat d_x|$ & 0.25, 3\\
Unscented transform $\alpha_{\mathrm{UT}}$, $\beta_{\mathrm{UT}}$, $\kappa_{\mathrm{UT}}$ & 0.5, 2, 0\\
Lateral sub-steps $N_s$ of the UKF process model & 3\\
\bottomrule
\end{tabularx}

\end{table}

\FloatBarrier  %
{\footnotesize\linespread{0.96}\selectfont
\setlength{\bibsep}{0pt plus 0.3pt}
\bibliographystyle{IEEEtranN}
\bibliography{ref}
}

\end{document}